\documentclass[manuscript,screen]{acmart}

\AtBeginDocument{%
  \providecommand\BibTeX{{%
    \normalfont B\kern-0.5em{\scshape i\kern-0.25em b}\kern-0.8em\TeX}}}

\setcopyright{rightsretained}
\copyrightyear{2022}
\acmYear{2022}
\acmDOI{} 

\acmConference[Technical Report]{ArXiv}{November 10,
  2022}{New York, NY}

\usepackage{xargs} 
\usepackage{xspace}
\usepackage[prologue,pdftex,dvipsnames]{xcolor}  
\usepackage[colorinlistoftodos,prependcaption,textsize=tiny]{todonotes}
\usepackage{tabularx}
\usepackage{colortbl}
\usepackage{xcolor}
\definecolor{lightgray}{gray}{0.8}

\usepackage{multirow}
\usepackage{hyperref}

\usepackage{pifont}
\newcommand{\cmark}{\ding{51}}%
\newcommand{\xmark}{\ding{55}}%

\makeatletter
\DeclareRobustCommand\onedot{\futurelet\@let@token\@onedot}
\def\@onedot{\ifx\@let@token.\else.\null\fi\xspace}

\def\eg{\emph{e.g}\onedot} 
\def\ie{\emph{i.e}\onedot} 
 
\def\etc{\emph{etc}\onedot} 
 
\def\etal{\emph{et al}\onedot}
\makeatother

\begin{document}

\title[Casual Conversations v2] {Casual Conversations v2: Designing a large consent-driven dataset to measure algorithmic bias and robustness}

\author{Caner Hazirbas}
\affiliation{%
  \institution{Meta AI}
  \city{New York}
  \country{USA}}
\email{hazirbas@meta.com}

\author{Yejin Bang}
\authornote{Both authors contributed equally to this research.}
\email{yjbang@connect.ust.hk}
\author{Tiezheng Yu}
\authornotemark[1]
\email{tyuah@connect.ust.hk}
\affiliation{%
  \institution{Hong Kong University of Science and Technology}
  \country{Hong Kong}
}

\author{Parisa Assar}
\affiliation{%
  \institution{Meta AI}
  \city{Menlo Park}
  \country{USA}
}
\email{parisar@meta.com}

\author{Bilal Porgali}
\affiliation{%
 \institution{Meta AI}
 \city{London}
 \country{United Kingdom}}
\email{porgali@meta.com}

\author{Vítor Albiero}
\affiliation{%
 \institution{Meta AI}
 \city{Menlo Park}
 \country{USA}}
\email{valbiero@meta.com}

\author{Stefan Hermanek}
\affiliation{%
 \institution{Meta AI}
 \city{New York}
 \country{USA}}
\email{sjhermanek@meta.com}

\author{Jacqueline Pan}
\affiliation{%
  \institution{Meta AI}
  \city{Menlo Park}
  \country{USA}}
\email{jackiepan@meta.com}

\author{Emily McReynolds}
\affiliation{%
  \institution{Meta AI}
  \city{Seattle}
  \country{USA}}
\email{emcr@meta.com}

\author{Miranda Bogen}
\affiliation{%
  \institution{Meta AI}
  \city{Washington, DC}
  \country{USA}}
\email{mbogen@meta.com}

\author{Pascale Fung}
\affiliation{%
  \institution{Hong Kong University of Science and Technology}
  \city{Hong Kong}
  \country{Hong Kong}}
\email{pascalefung@meta.com}

\author{Cristian Canton Ferrer}
\affiliation{%
  \institution{Meta AI}
  \city{Seattle}
  \country{USA}}
\email{ccanton@meta.com}

\renewcommand{\shortauthors}{Hazirbas, et al.}

\begin{abstract}
Developing robust and fair AI systems require datasets with comprehensive set of labels that can help ensure the validity and legitimacy of relevant measurements. Recent efforts, therefore, focus on collecting person-related datasets that have carefully selected labels, including sensitive characteristics, and consent forms in place to use those attributes for model testing and development. Responsible data collection involves several stages, including but not limited to determining use-case scenarios, selecting categories (annotations) such that the data are fit for the purpose of measuring algorithmic bias for subgroups and most importantly ensure that the selected categories/subcategories are robust to regional diversities and inclusive of as many subgroups as possible.
  
Meta, in a continuation of our efforts to measure AI algorithmic bias and robustness\footnote{\url{https://ai.facebook.com/blog/shedding-light-on-fairness-in-ai-with-a-new-data-set}}, is working on collecting a large consent-driven dataset with a comprehensive list of categories. This paper describes our proposed design of such categories and subcategories for \textit{Casual Conversations v2}.
\end{abstract}



\maketitle

\section{Introduction}
There is no shortage of literature~\cite{USreadout22, TianImage22, mehrabi2021survey} on AI systems performing inconsistently across different sub-groups of society, and scholars have continued to advocate for further diversity of such datasets. Several recent studies~\cite{lahoti2020fairness, han2022towards, du2021fairness, jung2022learning, amini2019uncovering} propose various learning strategies for AI models to be well-calibrated across all protected subgroups, while others focus on collecting responsible datasets~\cite{hazirbas2022towards, liu2022towards, smith2022m} to make sure evaluations of AI models are accurate and algorithmic bias can be measured while promoting data privacy.

There has been much criticism regarding the design choice of the publicly used datasets, such as for ImageNet~\cite{deng2009imagenet, khan23subjects, Crawford2021, hanley2020ethical}. Discussions are mostly focused on concerns around collecting sensitive data about people without their consent. Casual Conversations v1~\cite{hazirbas2022towards} was one of the first benchmarks that was designed with permission from participants. However, that dataset has several limitations: samples were collected only in the US, the gender label is limited to three options, and only age and gender labels are self-provided with the permission of the participants. While it has enabled many researchers~\cite{dooley2022robustness, goyal2022fairness} to evaluate their models across these and other dimensions, it does not cover many categories that may be needed to more fully evaluate models such as physical attributes and more granular subcategories.

In order to understand the landscape of a responsible dataset for model evaluation in audio/vision/speech models, we conducted on a literature survey that helped us understand how we should identify new self-provided label options and how those labels should be sub-categorized in order to be more inclusive. The objective of the design is to provide a list of categories to measure potential algorithmic failure on as many use-cases as possible,~\eg, computer vision, audio, speech and language models. Therefore to facilitate this, we decided to collect ten categories. Many of the recent works~\cite{hazirbas2022towards, liu2022towards, goyal2022fairness, Raji19, Grother2018, Cook2019, Howard2019, mengesha_i_2021, mehrabi2021survey, chen2022exploring, krishnapriya2020issues, albiero2020analysis, albiero2020does, albiero2021gendered} break down their analysis mostly based on age, gender, race, language, geo-location, apparent skin-tone, and voice timbre. However, there are several other categories that have not been covered in most of the datasets but will improve the inclusion of less represented sub-groups of people and circumstances such as physical attributes (body size, body/face markings, tattoo,~\etc), and activity (standing, walking,~\etc).

While we do believe demography-related labels are highly salient to measuring bias, we also add a set of categories that are relevant to model performance for the relevant use-cases and thus can help measure robustness of the models that may directly impact or be experienced as model fairness gaps, such as different recording setups (lighting, camera quality, environment,~\etc.). 

In Section~\ref{sec:categories} we provide our literature review for each category and present proposed subcategories of each that will be included in our dataset. Further, in Section~\ref{sec:excluded} we discuss other categories that are commonly used for measuring algorithmic bias, \eg, race, and elaborate the reasons why we opted not to solicit them in our data collection.


\section{Proposed Categories\label{sec:categories} in the Casual Conversations v2 dataset}
In the following section, we discuss each of these categories, provide a list of proposed subcategories and further evaluate their utility for algorithmic bias measurement. In Table~\ref{tab:categories}, we list all categories and indicate which ones will be self-provided and which ones will be annotated.

\begin{table}[ht]
    \arrayrulecolor{lightgray}
    \begin{tabular}{ll|cc}
    \textbf{Section} & \textbf{Category} & \textbf{Self-provided} & \textbf{Annotated} \\
    \toprule
    Section~\ref{sec:age} & Age & \cmark & \\
    Section~\ref{sec:gender} & Gender & \cmark & \\
    Section~\ref{sec:language} & Language/Dialect & \cmark & \\
    Section~\ref{sec:location} & Geo-Location (Country/State/City) & \cmark & \\
    Section~\ref{sec:disability} & Disability & \cmark & \\
    Section~\ref{sec:physical} & Physical Adornments and Physical Attributes & \cmark & \\
    Section~\ref{sec:voice} & Voice Timbre & & \cmark \\
    Section~\ref{sec:skin_tone} & Apparent Skin Tone & & \cmark \\
    Section~\ref{sec:recording} & Different Recording Setup & & \cmark \\
    Section~\ref{sec:activity} & Activity & & \cmark \\
    & & & \\
    \textbf{Section} & \textbf{Excluded Categories} \\
    \midrule
    Section~\ref{sec:excluded} & Race \& Ethnicity & \xmark & \xmark \\
    Section~\ref{sec:excluded} & Facial Expressions & \xmark & \xmark \\
    \bottomrule
    \end{tabular}
    \caption{List of proposed self-provided and annotated categories that will present in the \textit{CCv2} dataset. We exclude several categories in this dataset and provide reasons in Section~\ref{sec:excluded}.}
    \label{tab:categories}
\end{table}

\subsection{Age\label{sec:age}} 
Investigating the robustness of AI models for age groups is important in building fair AI algorithms. Notably, many current public datasets~\cite{kumar2009attribute, hazirbas2022towards} perform analysis on the various age groups. However the definition of these age groups vary from dataset to dataset and we would like to propose the categorization shown in Table~\ref{tab:age_groups}. This proposal summarizes how we plan to categorize age, which is synthesized from various governmental and academic resources. Though fairness for AI with regard to youth is a priority for policymakers, we propose to collect data from people who are older than the age of maturity, which varies by country between ($18-22$) given equally salient privacy concerns when it comes to younger populations~\cite{AIforChildren}.

\paragraph{\textbf{Governmental resources.}} Many government population statistics include age distribution. Here we consult the statistical standards of age groups from multiple governmental resources. Both Singapore Department of Statistics~\cite{SingaporeAge} and World Bank~\cite{USAge} divide the age groups into $0-14$, $15-64$, and $65+$. Statistics Canada~\cite{CanadaAge} chooses to divide the age groups into Children ($0-14$), Youth ($15-24$), Adults ($25-64$) and Seniors ($65+$). In addition, United Nations~\cite{UNAge} uses $0-14$, $15-24$, $25-44$, $45-64$ and $65+$ age groups. 

\paragraph{\textbf{Scientific resources.}} We also reviewed the current public datasets. Many works have been performed on using face representation to predict age groups~\cite{huang2017age,dong2016automatic,agbo2021deep,duan2018hybrid}. For example, an age group classification dataset based on facial features~\cite{horng2001classification} divides the groups into Baby ($0-2$), Young Adults ($3-39$), Middle-age Adults ($40-59$), and Old Adults ($60+$). An Automatic Speech Recognition dataset that aims to quantify age bias~\cite{feng2021quantifying} introduces age groups includes children ($7-11$), Teenagers ($12-16$), Adults ($17-64$) and Older Adults ($65+$). Casual Conversations v1 dataset~\cite{hazirbas2022towards} groups the category for age into three buckets that are $18-30$, $31-45$ and $46-85$. FairFace: a face attribute dataset~\cite{karkkainen2021fairface} and Common Voice: a speech corpus~\cite{ardila2020common} use age buckets mainly at ten-year intervals. 

\begin{table}[!h]
    \centering
    \begin{tabular}{l|c}
    \toprule
    Age Groups        & Age Intervals \\ \midrule
    Young Adults      & $X$-24 \\
    Adults            & 25-40 \\
    Middle-age Adults & 41-65 \\ 
    Old Adults        & 65+ \\ \bottomrule
    \end{tabular}
    \caption{The proposed age intervals for age groups. $X$, the age of maturity varies by country between $18-22$.}
    \label{tab:age_groups}
\end{table}

\subsection{Gender\label{sec:gender}} 
Gender is one of the most actively discussed protected attributes when it comes to algorithmic bias and robustness of AI systems~\cite{yao2017beyond, buolamwini2018gender, sun2019mitigating, wang2019balanced, wang2020towards}. Until recent years and even today in many cultures, gender identities have often been interchangeably confused with sex at birth (male and female). Thus, cumulative data or data annotation in much research considers gender as binary (man and woman). Most of the facial recognition research~\cite{Raji19, Grother2018, Cook2019, Howard2019} only considers binary genders in their analyses. However, gender is a social construction that changes over time and differs among various cultures~\cite{who_gender}. Thus, choosing subcategories in gender requires a thorough assessment.

There is no fixed number of subgroups for gender identity; rather, it is more of a spectrum than a discrete matter~\cite{brito_2018}. There are more than \textit{twenty five} gender subcategories according to the LGBTQIA Resource Center Glossary~\cite{lgbtqia_center_2022}. However, as stated in~\cite{jordan_2021} (guideline for gender options in the surveys), an exhaustive list with every possible option may overwhelm and cause fatigue to the participants. In order to ensure inclusion of all possible gender identities, we recommend offering "\textit{Not stated above, please specify}`` option where participants can indicate their genders in a free-form text. The following provides the overview of gender categorization in perspectives from governmental, scientific and survey resources.

\paragraph{\textbf{Governmental resources.}} 
Conventionally, many governments have adopted heteronormative binary gender,~\ie, ``\textit{Cis Man/Male (M)}'' and ``\textit{Cis Woman/Female (F)}'', as a common choice for legal and institutional systems and official documents. However, with increased awareness of ``\textit{Non-binary}'' genders in past few years, there has been legal recognition with an option of ``\textit{Non-binary}''~\cite{travelstategov, us_eeoc_2022, gov_uk_2019}. Seventeen countries around the world legally recognize the \textit{Non-binary} gender option on their passports (\ie, gender-neutral passport), including Argentina, Austria, Australia, Canada, Colombia, Denmark, Germany, Iceland, India, Ireland, Malta, Nepal, the Netherlands, New Zealand, Pakistan and the United States\footnote{The data are as of September 2022.}~\cite{the_economist_2022}. Besides, there is sometimes an option not to disclose their gender identity (\eg, ``\textit{Preferred not to say}'') in some governmental survey or consensus, respecting one's willingness for the disclosure. However, further specification about the \textit{Non-binary} genders is still very limited. 

In the United States, many official documents now require the option of ``\textit{Non-binary}'' as an effort to promote greater equity and inclusion. This includes adding ``\textit{Non-binary}'' as a gender marker printed on the U.S. passport (starting from 2023)~\cite{travelstategov} and in the Equal Employment Opportunity Commission (EEOC)'s charge intake process~\cite{us_eeoc_2022}. A survey of the LGBTQ+ population by the UK government~\cite{gov_uk_2019} has given more options including ``\textit{Trans Man/Woman}'', ``\textit{I do not want to say}'', ``\textit{I do not know}'' and ``\textit{Something else}''. The ``\textit{Something else}'' option allows participants of the survey to indicate their gender by themselves. The Australian government releases a standard for sex, gender, variations of sex characteristics and sexual orientation variables to standardize the collection and dissemination of related data~\cite{aus_stat}. Table \ref{tab:gender} shows the considered or required gender options for aforementioned consensus or surveys.

\begin{table}[ht!]
\resizebox{\linewidth}{!}{
    \arrayrulecolor{lightgray}
    \begin{tabular}{l|ccccc}
        \toprule
        Considered Genders & UK National LGBT Survey~\cite{gov_uk_2019} & Gender Survey Suggestion~\cite{jordan_2021} & U.S. Passport~\cite{travelstategov} & U.S. EEOC~\cite{us_eeoc_2022} & Australian Govt.~\cite{aus_stat} \\
        
        \midrule
        Cis Woman & Woman/girl & Female & Female (F) & Female (F) & Female (Sex at birth) \\
        Cis Man & Man/Boy & Male & Male (M) & Male (M) & Male (Sex at birth) \\
        Trans Man & Trans Man & Transgender Male &  &  &  \\
        Trans Woman & Trans Woman & Transgender Female &  &  &  \\
        Non-binary &
        \begin{tabular}[c]{@{}l@{}}Non-binary/\\ Genderqueer/\\ Agender/\\ Gender-fluid\end{tabular} &  & \begin{tabular}[c]{@{}l@{}}Another gender\\ identity (X) \end{tabular} & Non-binary (X) & Non-binary
        \\
        \begin{tabular}[c]{@{}l@{}}{}Not stated above\\(free-form) \end{tabular} & I do not know &
        \begin{tabular}[c]{@{}l@{}}{}Gender Variant /\\ Not-Conforming\end{tabular} &  &  & \begin{tabular}[c]{@{}l@{}}{[}I/they{]} use a \\ different term \\ (please specify)
    \end{tabular} \\
 & Something else & Not Listed &  &  & \\
Prefer Not to Say & I do not want to say & Prefer Not to Answer &  &  & Prefer not to answer \\

\bottomrule

\end{tabular}}
\caption{Gender subcategories considered or/and required by different governmental resources and social perspectives. EEOC stands for the Equal Employment Opportunity Commission.
}
    \label{tab:gender}
\end{table}

\paragraph{\textbf{Scientific resources.}} 
Gender is one of the most important protected attributes discussed within the topics of AI fairness. Broadly speaking, there are two cases where gender categorization is required 1) perceived gender: labeling the data with different gender categories when the data already includes gendered information (\eg, utilizing gender pronouns for constructing data for gender-bias in text or labelling human figures for gender in images) 2) gender identity of participants: collecting the personal information from data construction contributors (\eg,  gender of actors in video data, gender of data annotators, gender of the authors of certain online comments). The former are cases where gender information has not been provided directly thus a third party assumes the gender information by observing already existing data. The latter, in contrast, is where gender information can be gathered from the participants or the data builder. We looked into both approaches to gender categorizations in scientific resources.

Although there has been increased awareness regarding gender inclusiveness in the AI research community, binary gender is still a common categorization choice of gender attribute~\cite{park2018reducing, feng2021quantifying, ma2020powertransformer, lee2019understanding, nadeem2021stereoset, solak_2019}. For instance, there are only a few exceptions incorporating ``\textit{Non-binary}'' genders despite of the extensive list of research on  gender bias mitigation in NLP community in past few years~\cite{sun2019mitigating}. One of the reasons for considering binary genders is the difficulty of data collection with ``\textit{Non-binary}'' genders. For example, gendered pronouns are clearer for binary genders (\eg,  she, he) when crawling data from web-data in the NLP research. As a result, there is scarcity of such data that includes ``\textit{Non-binary}'' genders (nonbinary people may also be referred to with pronouns they do not prefer in written context, and a common pronoun, “they,” also more commonly refers to groups of people). As an effort for inclusion,~\cite{wang2019balanced, yao2017beyond, ma2020powertransformer, wang2020towards} showed understanding of ``\textit{Non-binary}'' genders and gender fluidity, but mostly limited to the acknowledgement. Some scholars go further and provide rationale on why ``\textit{Non-binary}'' genders are not included to compensate their limited inclusion of gender diversity. 

``\textit{Non-binary}'' genders are included when the gender of the data building contributors can be explicitly asked. Maarten~\etal~\cite{sap2020social} included ``\textit{Non-binary}'' genders option for collecting the personal information of data annotators and commonly used speech data set Common Voice~\cite{ardila2020common} extended the options to have ``\textit{Other}''.

\paragraph{\textbf{Research Guideline/Survey Design Guideline}} While  the topic of inclusion and diversity regarding gender has recently received more attention in AI and machine learning research, other research fields have extensive explorations that can be brought into AI/ML research. There are inclusive language guidelines for asking gender when collecting data from an individual from various institutions or organizations~\cite{jordan_2021}. 
In addition to the binary genders (Woman and Man), most commonly recommended options are ``\textit{Non-binary}'' and ``\textit{Non-disclosure}'' option (\eg, ``\textit{Prefer not to answer/say/respond/disclose}''). Many of the guidelines also recommend to include ``\textit{Transgender}(s)'' as an option besides ``\textit{Non-binary}'',~\eg, some include it as ``\textit{Transgender}''~\cite{csusm} while some further separate ``\textit{Trans Man}'' and ``\textit{Trans Woman}''~\cite{utah_2019}. An option of additional response (self-indication) is also recommended by most of the guidelines. To provide the self-indication option, it is recommended to use a term of  ``\textit{Not Listed/Referred}'' to avoid any negative connotation of the term ``\textit{Other}'' which is a common practice for indicating the not-listed options~\cite{jordan_2021}.

\paragraph{\textbf{Proposed final list of subcategories.}} While we avoid exhaustion-of-choices and survey fatigue, we came to a final list of subcategories as follow: \textit{Cis Man}, \textit{Cis Woman}, \textit{Transgender Man}, \textit{Transgender Woman}, \textit{Non-binary}, \textit{Not stated above, please specify (free-form)}, \textit{Preferred not to say}. We also provide definitions of these gender terms to the participants during data collection:

\begin{itemize}
    \item \textit{Cisgender}: an individual whose gender identity aligns with those typically associated with the biological sex assigned to them at birth. 
    \item {\textit{Transgender} (also shortened to trans) refers to an individual whose gender identity and/or expression is different from cultural expectations based on the sex they were assigned at birth. Being transgender does not imply any sexual orientation.
    \begin{itemize}
        \item \textit{Transgender man}: an individual whose biological sex was assigned female at birth, but identifies as a man.
        \item \textit{Transgender woman}: an individual whose biological sex was assigned male at     birth, but identifies as a woman.
    \end{itemize}
    }
    \item \textit{Non-binary}: a term used to describe a person who does not identify exclusively as a man or a woman. Non-binary people can identify as being both a man and a woman, somewhere in between, or fall completely outside the gender binary. Non-binary can also be used as an umbrella term encompassing identities such as genderqueer or gender-fluid.
\end{itemize}

\subsection{Language/Dialect\label{sec:language}} 
Linguists have long established categories and subcategories for languages and dialects. There is likely to be no need for AI researchers to re-define this categorization for languages. Rather, it is important to understand which languages to include when constructing a multi-lingual dataset. We conducted our literature review in two directions, \ie, 1) most commonly used languages and 2) a more diverse set of languages that takes into account low resource languages~\cite{wrro152844}.

When we enriched the selected languages/dialects to a more diverse set, we gathered more than a \textit{hundred} subcategories. In order to find the right balance, we first listed the most commonly spoken/used \textit{nineteen} languages and extended them to a more diverse set with several other languages. These \textit{nineteen} languages include English, Spanish, Japanese, Mandarin/Cantonese/Wu (Shanghainese), Korean, Vietnamese, Hindi/Tamil/Telugu, Arabic, Farsi, French, German, Italian, Slavic Languages, Portuguese, Turkish, Swahili, Thai, Bahasa (Indonesian) and American Sign Language (ASL). The additional more commonly used languages are Dutch, Greek, Swedish, Finnish, Hungarian, Filipino, Hebrew, and Bengali.  

Languages also have linguistic variations (\ie, dialects) that differ in grammar, pronunciation, and/or vocabulary~\cite{trudgill2017dialects, trousdale2010introduction}. In practice of collecting a dataset, dialects are often not clearly indicated, either because they are not considered or in some cases are included as optional. In the online crowd-sourced speech data Common Voice~\cite{ardila2020common} dataset, volunteers indicate their dialects from a predefined list (\eg,  the United States English, Australian English, the United Kingdom English) or self-declare. We recommend letting participants provide their dialects in a free-form text box during data collection. For instance, a participant may state their spoken language as Arabic while they also provide their dialect as Levantine Arabic.

\begin{table}[ht!]
\resizebox{\linewidth}{!}{
    \arrayrulecolor{lightgray}
    \begin{tabular}{lcccccc}
        \toprule
        \begin{tabular}[c]{@{}c@{}}Most commonly \\used languages \end{tabular} & 
        \begin{tabular}[c]{@{}c@{}}World most \\ Top-20 spoken \\ languages \\\cite{eberhard_simons_fenning_2022}\end{tabular} & \begin{tabular}[c]{@{}c@{}}Top-30 spoken \\ languages at \\ home in US$^\ast$ \\~\cite{census_bureau}\end{tabular} & \begin{tabular}[c]{@{}c@{}}World's most \\ used Top-10 \\ languages \\ in the Web\\\cite{internet_world_stats}\end{tabular} & \begin{tabular}[c]{@{}c@{}}Top-30 languages\\ sorted by number \\ of Wikipedia \\articles\\\cite{wikipedia_2022}\end{tabular} & \begin{tabular}[c]{@{}c@{}}UN official \\ languages\\\cite{un_official_lang}\end{tabular} & \begin{tabular}[c]{@{}c@{}} Top-10 languages\\studied around \\the world\\from Duolingo\\~\cite{blanco_2021}\end{tabular}\\
        \midrule
        English & \checkmark &  & \checkmark & \checkmark & \checkmark & \checkmark \\
        \begin{tabular}[c]{@{}l@{}}Spanish \\ (Castilian \\ and Neutral)\end{tabular} & \checkmark & \checkmark & \checkmark & \checkmark & \checkmark & \checkmark \\
        Japanese & \checkmark & \checkmark & \checkmark & \checkmark &  & \checkmark \\
        \begin{tabular}[c]{@{}l@{}}Mandarin/\\ Cantonese/\\ Wu (Shanghainese)\end{tabular} & \checkmark & \checkmark & \checkmark & \checkmark & \checkmark & \checkmark \\
        Korean &  & \checkmark &  & \checkmark &  & \checkmark \\
        Vietnamese & \checkmark & \checkmark &  & \checkmark &  &  \\
        Hindi, Tamil, Telugu & \checkmark & \checkmark &  &  &  &  \\
        Arabic & \checkmark & \checkmark & \checkmark & \checkmark & \checkmark &  \\
        Farsi &  &  &  &  &  &  \\
        French & \checkmark & \checkmark & \checkmark & \checkmark & \checkmark & \checkmark \\
        German & \checkmark & \checkmark & \checkmark & \checkmark &  & \checkmark \\
        Italian &  & \checkmark &  & \checkmark &  & \checkmark \\
        Slavic languages & \checkmark & \checkmark & \checkmark & \checkmark & \checkmark & \checkmark \\
        Portuguese & \checkmark & \checkmark & \checkmark & \checkmark &  & \checkmark \\
        Turkish & \checkmark &  &  & \checkmark &  &  \\
        Swahili &  &  &  &  &  &  \\
        Thai &  & \checkmark &  &  &  &  \\
        Bahasa (Indonesian) & \checkmark &  & \checkmark & \checkmark &  &  \\
        American Sign Language &  &  &  &  &  & \\
        \bottomrule
    \end{tabular}}
\caption{Most commonly used languages based on various resources. $^\ast$This statistics excludes English.}
\label{tab:most_common_lang}
\end{table}

\paragraph{\textbf{Most Commonly Used Languages.}} 
For our dataset, we took a holistic view and while we took into consideration the statistics of the most commonly used languages. We also reviewed various sources that demonstrate a compilation of the frequently used languages including Duolingo\footnote{An online language learning platform; It is the world's most-downloaded education app.; https://www.duolingo.com}, Web languages, Wikipedia articles (Table \ref{tab:most_common_lang}). We first checked the most spoken languages world-wide (most Top-20 spoken languages)~\cite{eberhard_simons_fenning_2022} as well as nation-wide in the U.S. (Top-30 Languages spoken at home other than English in the U.S.)~\cite{census_bureau}. Besides the spoken languages, the most commonly used languages online are referred through statistics on World's most used Top-10 languages in the Web (2020)~\cite{internet_world_stats} and Top-30 languages sorted by the number of Wikipedia articles~\cite{wikipedia_2022}. Other references include the U.N. official languages~\cite{un_official_lang} and Top-10 languages studied around the World (2021) from the Duolingo app~\cite{blanco_2021}. As a result, the list provided in the Table~\ref{tab:most_common_lang} below is composed by counting the intersections of the listed sources as well as the selected languages from the research papers as follow.

\paragraph{\textbf{Diverse set of Languages.}} We explored the diverse set of languages by referring to the literature in the field of NLP. Language understanding has always been an important task in the development of the NLP models. Since the 1990s, much research effort has been put into solving multilingual problems~\cite{liu2013asgard,litvak2010new,nothman2013learning}. Others have provided  multilingual datasets. Meta AI proposed a Multilingual LibriSpeech (MLS) dataset, which includes \textit{eight} languages' speech data from audio books~\cite{pratap2020mls}. Google Research introduced a multilingual dataset that consists of more than \textit{forty} languages, in total around \textit{forty billion} characters~\cite{guo2020wiki}. Many multilingual benchmarks, such as XGLUE~\cite{liang2020xglue} and XTREME~\cite{hu2020xtreme} have been introduced to evaluate the multilingual AI models. With the development of large-scale pre-training, pre-trained multilingual models (\eg,  mBART~\cite{liu2020multilingual} and mT5~\cite{xue2020mt5}) show significant success in immensely complicated tasks. In addition, Meta AI has built the first unified model for the multilingual machine translation that translates directly between the \textit{two hundred} languages without relying on the English data~\cite{fan2021beyond}.

\subsection{Geo-Location (Country/State/City)\label{sec:location}} 
As many open-source datasets are collected via crowd-sourcing, the geographical location of the speech/video data recording is not often shared. The well-known public speech datasets, Common Voice~\cite{ardila2020common}, M-AILABS Speech dataset~\cite{solak_2019} and LibriSpeech~\cite{panayotov2015librispeech}, also do not include any geographical information but only assume the speaker of crowd sourced data is a native speaker. The Casual Conversations v1 dataset is one of the few examples that shares geo-location information where videos were recorded but it is only limited to \textit{five} US cities.

However, geo-location can be useful information in speech and natural language data to assess the models' robustness. For instance, as one of the contributors of linguistic variations (dialect)~\cite{beal2010introduction}, geo-location is used to analyze AI speech problems for the same country but for different regions and/or cities. Thus, we propose instead to collect geo-location (where videos were collected) in addition to the  language spoken in each video. While granularity of geo-location could be as precise as longitude/latitude, we believe country, state/region and city will be enough to assess model robustness and fairness with respect to language and dialects.


\subsection{Disability\label{sec:disability}}  
Building AI systems that can better serve people with disabilities has been a major aspiration within the research community~\cite{whittaker2019disability,bennett2020point}. The key theme of prominent ``Nothing about us without us'' movement~\cite{Charlton1998Nothing}, casts light on the criticality of active inclusion of disabled participants.  However, approaches for evaluating the AI models' effectiveness for disabled people is not extensively investigated in comparison to other protected characteristics such as age and gender in audio and vision systems.~\cite{DBLP:journals/corr/abs-1811-10670} points out the diversity and outliers as one of the main challenges for AI fairness for people with disabilities as the disability is not \textit{a simple variable with a small number of discrete values} unlike other attributes such as age. One individual may have several different disabilities with varying degrees, and ability may evolve over time -- it requires the consideration of degree and diversity.

Since our dataset does not focus on specific disability, we considered the criteria for disability classification from different perspectives. Disability is generally classified into physical, sensory, intellectual and mental~\cite{DisabledWorld, TypesofDisabilities} categories. In academic research works, there has been no clear set of categories. The category of disability itself is generally considered for assessing bias towards the people with disabilities without any subcategories labels in~\cite{nangia2020crows}.~\citep{smith2022m} approached with few more categories of auditory, neurological, visual and unspecific. Academic researchers have also divided disability into hierarchical categories from general to specific subgroups ~\cite{graham2019patterns}. 

We propose disability categories as the following:  vision (blind, low-vision, visual field loss, vision-impaired, color-blindness, loss of depth perception, glasses-wearing); hearing (deaf, hard-of-hearing, hearing with one ear, cochlear-implant-using), physical (mobility, dexterity); speech (aphasic, speech-impaired, speech loss), cognitive (Learning/Neuro-diversity); use of assistive technology.

\begin{table}
\begin{tabular}{p{3cm}p{11cm}}
\toprule
 & Subcategories \\ \midrule
Physical Adornments & Have hair cover, Hair color (dye), Have beard/mustache, Have face covering, Have face mask, Have make-up, Have eye wear (sunglasses, glasses, contact lenses,~\etc), Have ear wear (ear rings), Have visible tattoos, Have bindi (South Asian face markings), Have visible piercings \\\midrule
Physical Attributes & Hair type, Hair color (original), Hearing aids, Eye color, Have birthmarks/fleckers/moles, Have other non-tattoo facial markings \\ \bottomrule
\end{tabular}
\caption{Proposed subcategories for Physical Adornments and Attributes.\label{tab:physical_attributes}}
\end{table}

\subsection{Physical Adornments and Physical Attributes\label{sec:physical}} 
Body and/or face markings are rarely taken into consideration in data collection as the collection of such data often requires more time and is a resource-intensive process. There is limited research work with physical attribute subcategories. Instead, we referred to the use cases of metaverse or online character creation such as Memoji (Apple)~\cite{apple_2022} and Avatar(Meta)~\cite{fb_avatar}. We propose to divide the category into physical adornments and physical attributes as illustrated in Table \ref{tab:physical_attributes}. For each participant, we suggest asking as YES or NO for each criteria, and provide an option of free-form text for participants to provide details if they wish.

\subsection{Voice Timbre\label{sec:voice}} 
Voice timbre is a unique tonal quality that gives color and personality to a person's voice. It is crucial to help make AI systems, especially Automatic Speech Recognition (ASR) models, more equally performant with regards to voice timbre. Voice timbre is generally categorized into Soprano, Mezzo Soprano, Alto, Contralto, Countertenor, Tenor, Baritone and Bass based on music theory~\cite{MusicToYourHome,MasterClassTimbre,RamseyTimbre}. Moreover, one special category for pre-pubescent voice is called Treble. Although we do not plan to include minors in our dataset, we intend to include this voice timbre label as well, as age may not be dispositive of voice timbre 

In practice, voice timbre is often used as a technical term for music theory, thus, it is hard for general individuals to self-identify their own voice timbres.
Academic publications either categorize voice from high-pitched to deep (\eg,  High-pitched, Ethereal, Sweet, Powerful, Gravelly and Deep)~\cite{sha2013singing}, or use gender/age-referring phrases such as middle-aged women, young men, or little girls~\cite{lyu2020convolutional}. One study utilizes a dataset that is collected from singers and separates three groups for both male and female~\cite{roers2009voice} (see Table~\ref{tab:Voice_Timbre}).

We propose three subcategories of low pitch, average pitch, and high pitch to ensure easier labeling. These subcategories provide intuitive names while enabling us to gather meaningful enough information for testing model performance in voice timbre variation.

\begin{table}[h]
    \centering
    \resizebox{\linewidth}{!}{
    \begin{tabular}{ccccccc}
    \toprule
    Ours           &~\cite{MusicToYourHome}   &~\cite{MasterClassTimbre}   &~\cite{RamseyTimbre}    &~\cite{sha2013singing} &~\cite{lyu2020convolutional} &~\cite{roers2009voice} \\ \midrule
    high pitch     & Soprano                  & Soprano                    & Soprano                & High-pitched          & Middle-aged Women           & Soprano \\ 
    average pitch  & Mezzo                    & Mezzo-soprano              & Mezzo Soprano          & Ethereal              & queen-style ladies          & Mezzosoprano\\ 
    low pitch      & Alto                     & Contralto                  & Alto                   & Sweet                 & Young girls                 & Alto\\ 
          & Tenor                    & Tenor                      & Contralto              & Powerful              & Cute little girls           & Tenor\\  
                   & Bass                     & Baritone                   & Countertenor           & Gravelly              & Elderly                     & Baritone\\ 
                   &                          & Bass                       & Tenor                  & Deep                  & Middle-aged men             & Bass\\ 
                   &                          &                            & Baritone               &                       & Young men                   & \\
                   &                          &                            & Bass                   &                       & Cute little boys            & \\ \bottomrule
    \end{tabular}}
    \caption{Suggested voice timbre subcategories and categorization of voice timbre from music theory and academic research.}
    \label{tab:Voice_Timbre}
\end{table}

\subsection{Apparent Skin Tone\label{sec:skin_tone}} 
Skin tone is an important attribute of human appearance, with significant variation from pale to dark. Recently, AI systems, especially computer vision models, have become controversial over concerns about the potential bias of performance varying based on the skin tone~\cite{krishnapriya2020issues, lu2019experimental,buolamwini2018gender}. Therefore, we include skin tone as one of the categories to collect data for the purposes of measuring algorithmic bias.

Skin tone (or skin color in some literature~\cite{bar2009role, xu2022color}) does not have one clear universal measure even within the field of dermatology~\cite{merler2019diversity}. In computer vision, the Fitzpatrick scale~\cite{sachdeva2009fitzpatrick} is currently the most commonly used numerical classification schema for skin tone due to its simplicity and widespread use~\cite{buolamwini2018gender, muthukumar2018understanding, wilson2019predictive, 10.1145/3306618.3314243}. However, some literature points out that the use of the Fitzpatrick scale may be problematic as it is an unreliable estimator of the skin pigmentation~\cite{okoji2021equity}.
Recently, the Monk Skin Tone (MST) Scale~\cite{monk2014skin} was proposed in collaboration with Google which is going to be utilized in several Google products~\cite{GoogleACloserLook} including search and photos. In contrast to the 6-tone Fitzpatrick scale, Monk has 10 tones that favors darker and lighter skin tones equally. Many studies in different fields have used the MST Scale~\cite{foy2017shade,monk2015cost,louie2018representations}.

We suggest using both Fitzpatrick and Monk scales for apparent skin tone annotation. Using both scales will allow clearer comparison with previous works that use Fitzpatrick scale while also enabling measurement based on more balanced skin tone information collection through Monk scale. 


\subsection{Different Recording Setup\label{sec:recording}} 

To understand the models' robustness and fairness, the different recording setup is also an important element to consider as all environments in the videos cannot be uniform across different datasets -- and such variation may have an underlying relationship to factors relevant to understanding unfair bias such as income level. From simple consideration of resolutions of the videos (\eg, high-definition, low-definition, smartphones) to environment (\eg, outdoor, indoor, beach,~\etc), we suggest to consider varieties of recording setups as suggested in Table \ref{tab:recording_setup}.

\begin{table}[h]
    \centering
    \begin{tabular}{p{0.4\columnwidth}p{0.5\columnwidth}}
    \toprule
    \textbf{Recording setup}                     & \textbf{Notes} \\
    \multirow{3}{*}{Setup}
        & Subscripted text for speech, \\
        & Unsubscripted free–form speech, \\
        & Interview \\
    \midrule
    \multirow{3}{*}{Environment}
        & Indoor, Outdoor, Beach, Field, Mountain, \\
        & Office, Living room, Bedroom, Kitchen, \\
        & Street, Sky, Water,~\etc \\
    \midrule
    Diverse weather settings            & - \\
    \midrule
    Diversity in northern and southern hemisphere (sun position) & - \\
    \midrule
    \multirow{3}{*}{Types of video quality}
        & Web camera, Laptop camera, \\
        & Smartphone camera, Professional camera,~\etc \\
        & (technical details about the camera if possible) \\
    \midrule
    Background noise                    & - \\
    \midrule
    \multirow{2}{*}{Camera position}
        & Hand-held phone, Selfie, \\
        & Laptop, Photography studio,~\etc \\
    Object bounding boxes and/or segmentation           & - \\
    Lighting                                            & Perceptually dark or light \\ \bottomrule
    \end{tabular}
    \caption{Different recording setup.}
    \label{tab:recording_setup}
\end{table}

\subsection{Activity\label{sec:activity}} 
Types of human activities are broadly categorized as gestures, actions and interactions~\cite{aggarwal2011human}. The subcategories for each activity category can be fine-grained to 200+ activities~\cite{caba2015activitynet}. Since annotating a whole list of human activities can be innumerable, we mainly focused on the most overlapping list of activities and also the most applicable actions, gestures and appearance in our data collection setup shown in Table~\ref{tab:activity}. We mainly referred to computer vision research paper that focuses on daily human activity detection~\cite{DBLP:journals/corr/abs-1708-08989, bobick2001recognition, aggarwal2011human, ku2018classification, anguita2012human}.

\begin{table}[h]
    \begin{tabular}{p{0.4\columnwidth}l}
        \toprule
        \textbf{Activity} & \textbf{Subcategories}\\
        \multirow{5}{*}{Action}
            & Standing \\
            & Walking \\
            & Sitting \\
            & Laying \\
            & Waving \\
        \midrule
        \multirow{3}{*}{Gesture}
            & Streching body \\
            & Raising hand/leg \\
            & Moving head \\
        \midrule
        \multirow{4}{*}{Appearance}
            & Full body visible \\
            & Upper body visible \\
            & Lower body visible \\
            & Only head visible \\
        \bottomrule
    \end{tabular}
    \caption{List of activities. We group activities into \textit{action}, \textit{gesture} and \textit{appearance}.}
    \label{tab:activity}
\end{table}

\section{Deliberate exclusion of certain categories\label{sec:excluded}}
In the previous Section~\ref{sec:categories}, we went over each of the ten different categories and the related literature and rationale for final sub categories. However, there are some categories that cannot be generalized to the world population and may not be relevant to the aforementioned measurement or ML training tasks. In this section, we explain the reason why we will not include them in our dataset.\\

\noindent\textbf{Race~\& Ethnicity}. Both race and ethnicity are categories that are often discussed in fairness analysis of AI systems \cite{karkkainen2021fairface, Wang_2019_ICCV, chen2022exploring, Wang_2020_CVPR}. For instance, most studies in face recognition fairness have heavily relied on race/ethnicity~\cite{karkkainen2021fairface,Wang_2019_ICCV,das2018mitigating}. However, Roth~\cite{roth2016multiple} argues that there is no single dimension that identifies the person's ``true'' or ``correct'' race because race is being experienced as a result of conflicting dimensions including self-identification, perception by others in the society, person's belief on others' perception, racial self-classification, skin color, racial appearance and racial ancestry. By decoupling race into its diverse dimensions, Roth demonstrates that race is a social construct and discusses further to show racial fluidity and racial boundary change. Furthermore, race has ambiguities to be well-defined and generalized to all regions (i.e., race and ethnicity are defined differently in different countries). Barjubani~\etal ~\cite{Barbujani2011} also highlights that ``\textit{agreeing on a catalogue of human races has so far proved impossible}.''

Recently, a shift towards using skin tone instead is starting to happen~\cite{Cook2019, krishnapriya2020issues, wang2021meta, buolamwini2018gender}, mostly due to the problems outlined here. Moreover, when evaluating visual or auditory models it is more accurate to consider actual visual attributes~\cite{FABBRIZZI2022103552} rather than race with its limitations and ambiguities. We already have location, apparent skin-tone and language/accent/dialect information which provide a robust baseline analysis of AI models performance. Thus, the additional collection of controversial/sensitive data about race/ethnicity fails to override the reasons not to.\\

\noindent\textbf{Facial Expression}. There has been a substantial amount of research conducted in the field of \textit{apparent emotion detection}~\cite{Aneja2018Learning, Livingstone2018Ryerson,Lucey2010IEEE, Kamachi1997Japanese,Sneddon2012IEEE, Singh2020ISafe, Langner2010Cognition, Ali2017Facial, Dimitrios2019Expression, li2017reliable, li2019reliable, Bagher2018Multimodal, AifantiNikiMUG, ZHAO2011Facial, albiero2018multi,batista2017aumpnet}. Nevertheless, the most recent studies have unearthed several risks of mis-contextualizing \textit{facial expressions}. Barrett~\etal~\cite{Barrett2019Emotional} discusses the challenges of inferring \textit{emotions} as  expressions (anger, fear, disgust, happiness,~\etc.) may be communicated in many different forms across different cultures and situations. As a result, mis-contextualizing of emotions may result in potential harms in the commercial applications. Hernandez~\etal~\cite{hernandez2021guidelines} provides \textit{twelve} guidelines for systematic assessment and reducing the risk presented by apparent emotion detection applications. Moreover, \textit{facial expression} could be potentially utilized to reveal unintended information of individuals~\cite{Hebbelstrup2022Using}. Considering all these risks and ambiguity in \textit{facial expression detection}, we decided to exclude this category from the list of labels we plan to collect for the dataset.

\section{Conclusion}
In this article, we presented a literature survey we have conducted for Casual Conversations v2,~\ie a large consent driven dataset designed for measuring algorithmic bias and robustness of AI systems. We proposed to collect \textit{six} self-provided (by the participants) and \textit{four} annotated categories, including \textit{age}, \textit{gender}, \textit{language/dialect}, \textit{geo-location (country/state/city)}, \textit{disability}, \textit{physical adornment}, \textit{physical attributes}, \textit{voice timbre}, \textit{apparent skin tone}, \textit{different recording setup} and \textit{activity}. For each of these categories, we provided a thorough review and discussed the proposed subcategories that we aim to collect. Furthermore, in the spirit of building a responsible dataset, three sensitive categories,~\ie \textit{race \& ethnicity} and \textit{facial expressions}, have been excluded for ethical and safety reasons that also aligns with our ongoing efforts and long term vision in Responsible AI.

This literature review has been carried out by Meta as part of our continued commitment to Responsible AI in collaboration with academic partners. \textit{Casual Conversations v2} will be unique in the academic world in terms of its proposed categories (which are rarely seen in the publicly available audio/vision benchmarks) and also the countries where the data collection will take place. Our dataset will enable the research community 1) to develop new models that are more fair, inclusive, and robust and also 2) to measure their models on these axes.

\begin{acks}
We would like to express our deepest gratitude to many partners involved in the process of this literature review and also in the construction of this dataset, including but not limited to Civil Rights, Accessibility, Responsible AI, AI Analytics, Assistant, Speech Recognition, and FAIR teams at Meta. We also want to thank Eric Smith, Skyler Wang, Tashrima Hossain, Sarah Smurthwaite, Matt Bonna and Ida Cheng for their support and contribution.
\end{acks}

\bibliographystyle{ACM-Reference-Format}
\bibliography{bibliography}

\end{document}